\documentclass{article}
\usepackage{spconf,amsmath,epsfig}
\usepackage[colorlinks, citecolor=green]{hyperref}
\usepackage{amssymb}
\usepackage{dsfont}
\usepackage{booktabs}
\usepackage{appendix}
\let\OLDthebibliography\thebibliography
\renewcommand\thebibliography[1]{
  \OLDthebibliography{#1}
  \setlength{\parskip}{0pt}
  \setlength{\itemsep}{0pt plus 0.3ex}
}
\newcommand{\ptitle}[1]{\noindent\textbf{#1}\hspace{5pt}}
\begin{document}\sloppy

\def\x{{\mathbf x}}
\def\L{{\cal L}}

\title{VRHCF: Cross-Source Point Cloud Registration via Voxel Representation and Hierarchical Correspondence Filtering}
%
\name{Guiyu Zhao, Zewen Du, Zhentao Guo, Hongbin Ma$^{*}$}
\address{National Key Lab of Autonomous Intelligent Unmanned Systems \\ 
School of Automation, Beijing Institute of Technology\\
$^{*}$Corresponding author, Email: mathmhb@bit.edu.cn}

\maketitle

\begin{abstract}
Addressing the challenges posed by the substantial gap in point 
cloud data collected from diverse sensors, achieving robust cross-source point 
cloud registration becomes a formidable task. In response, we present a novel 
framework for point cloud registration with broad applicability, suitable for 
both homologous and cross-source registration scenarios. To tackle the issues 
arising from different densities and distributions in cross-source point cloud 
data, we introduce a feature representation based on spherical voxels. 
Furthermore, addressing the challenge of numerous outliers and mismatches in 
cross-source registration, we propose a hierarchical correspondence 
filtering approach. This method progressively filters out mismatches, yielding 
a set of high-quality correspondences. Our method exhibits versatile applicability 
and excels in both traditional homologous registration and challenging 
cross-source registration scenarios. Specifically, in homologous registration 
using the 3DMatch dataset, we achieve the highest registration recall of 
95.1\% and an inlier ratio of 87.8\%. In cross-source point cloud registration, 
our method attains the best RR on the 3DCSR dataset, demonstrating a 9.3 percentage points 
improvement. The code is available at \url{https://github.com/GuiyuZhao/VRHCF}.
\end{abstract}
\section{Introduction}
\label{sec:intro}

Cross-source point cloud registration~\cite{huang2023cross} involves aligning point 
cloud data from different sensors and methods, such as SFM and Kinect or Kinect 
and LiDAR, to the same coordinate system through rotation and translation. 
This process holds promising applications in various domains, including 
unmanned driving, remote sensing, and medicine. 
We can leverage cross-source point cloud registration to facilitate data calibration, thereby integrating the strengths of various sensors.
This alignment facilitates multi-sensor fusion, opening avenues for advancements in various computer vision tasks.

\begin{figure}[t]
  \centering{\includegraphics[scale=0.4]{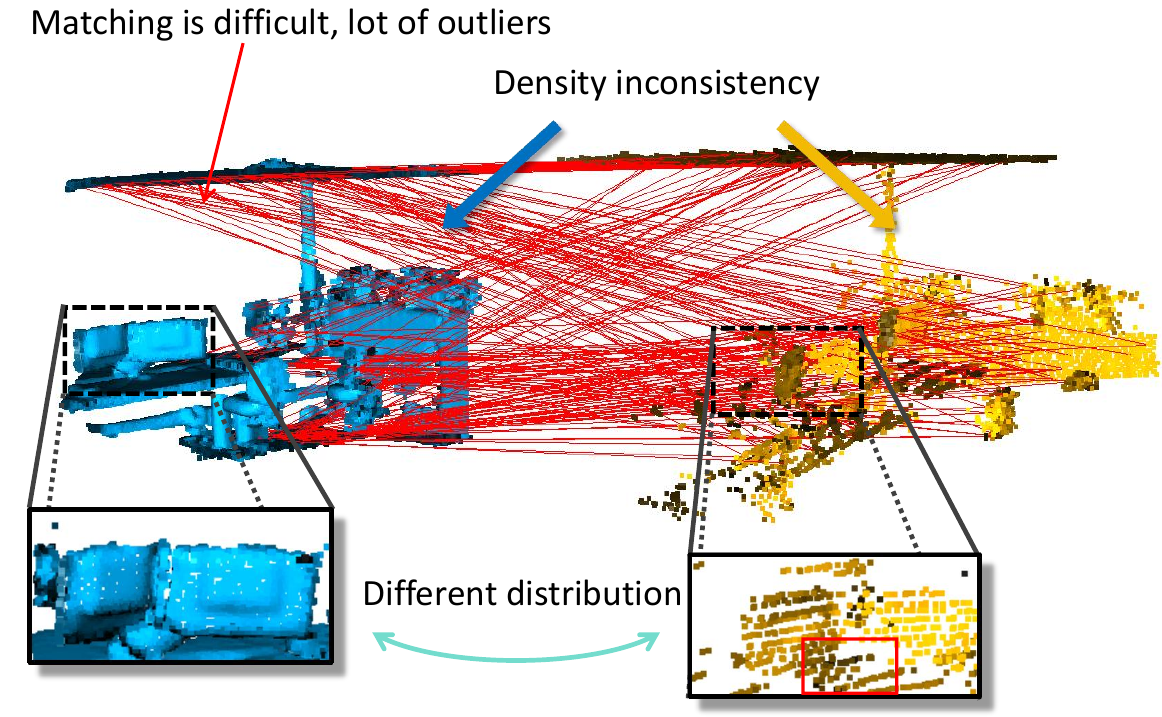}}  
  \vspace{-5pt}
  \caption{
    Challenges in cross-source point cloud registration.
      }
  \label{intro}

\end{figure}

Cross-source point cloud registration~\cite{huang2023cross} presents a formidable challenge 
within the broader task of point cloud registration. The inherent 
difficulty arises from two key challenges~\cite{huang2023cross}, as shown in Figure~\ref{intro}. 
\textbf{Firstly}, the diverse 
origins of point cloud data, stemming from different sensors, 
introduce significant disparities in cross-source datasets. 
Notably, the prominent issue is the density inconsistency and 
different distribution in point 
clouds, with LiDAR-generated point clouds exhibiting sparsity 
and line-like structures, in stark contrast to the dense nature 
of point clouds from depth cameras. This stark contrast complicates 
the task of learning similar features through feature metric 
learning. \textbf{Secondly}, the presence of numerous outliers in point 
clouds collected by distinct sensors, coupled with the intricate 
challenge of feature matching, results in a substantial number of 
mismatches in obtained correspondences. These dual challenges 
collectively pose substantial difficulties in achieving robust 
cross-source registration.

Currently, there is a lot of research focused on same-source 
registration~\cite{qin2022geometric,ao2021spinnet, huang2021predator, HybridPoint}, 
yet the majority of these approaches exhibit bad 
performance in cross-source registration. Even 
SpinNet~\cite{ao2021spinnet}, known for its strong generalization, falls short in 
achieving generalization across different sources, limiting its 
applicability. Simultaneously, numerous cross-source registration 
algorithms~\cite{huang2017coarse, zhao2023accurate, huang2019fast} 
have been introduced in recent years, primarily relying 
on optimization methods. Unfortunately, although these methods 
achieve some improvement in cross-source registration, 
they perform poorly in unseen scenes and noisy scenes.
In addition, most 
existing methods face challenges in effectively addressing the 
dual issues mentioned above, making it challenging to achieve 
robust cross-source registration.

Motivated by the challenges inherent in cross-source point cloud 
registration, we introduce a novel framework designed to 
address these difficulties, exhibiting good generalization ability across 
various scenes and excelling in both homologous and 
cross-source registration scenarios. \textbf{In response to the 
first challenge}, we present a feature representation based 
on spherical voxels. Departing from conventional downsampling 
methods~\cite{huang2017coarse} that risk feature loss, our approach leverages spherical 
voxels to achieve density-invariant representation without 
significant downsampling, thereby preventing the associated 
feature degradation. Feature refinement is subsequently 
conducted through the 3D cylindrical convolution network (3DCCN) backbone~\cite{ao2021spinnet} to derive density-invariant 
feature descriptors. \textbf{For the second challenge}, we introduce 
hierarchical correspondence filtering—a refined and 
progressive mechanism for filtering false matches, resulting 
in a high-quality set of correspondences. Finally, using SVD~\cite{ICP}, 
we achieve a more robust and accurate registration.

Our method demonstrates strong performance in both homologous 
and cross-source point cloud registration. Notably, in the 
cross-source 3DCSR \emph{Kinect-sfm} benchmark, our approach outperforms 
the original method, achieving a substantial 9.3-percentage-point 
improvement at 93.8\%, indicative of a more robust registration 
capability.
On the challenging \emph{Kinect-lidar} benchmark,
our method demonstrates a improvement of 1.2 percentage points (pp) and 3.9 pp in RR and FMR, respectively.
Additionally, in homologous point cloud registration, on the 
3DMatch dataset, our method outshines other strong baselines, 
achieving the best performance with 95.1\% RR and 87.8\% IR. Our key 
contributions include:

\begin{itemize}
  \setlength{\itemsep}{0pt}
  \setlength{\parsep}{0pt}
  \setlength{\parskip}{0pt}
  \item[$\bullet$] 
  We introduce a novel registration framework that exhibits robust and precise cross-source and homologous registration, 
  suitable for a diverse array of scenarios.
  \item[$\bullet$] A spherical voxel feature representation method is proposed to 
  realize the feature representation with consistent density and avoid the feature loss caused by downsampling.
  \item[$\bullet$]  hierarchical correspondence filtering is proposed to solve the problem of outliers and false matches in cross-source registration.
\end{itemize}

\begin{figure*}[t]
  \centering{\includegraphics[scale=0.5]{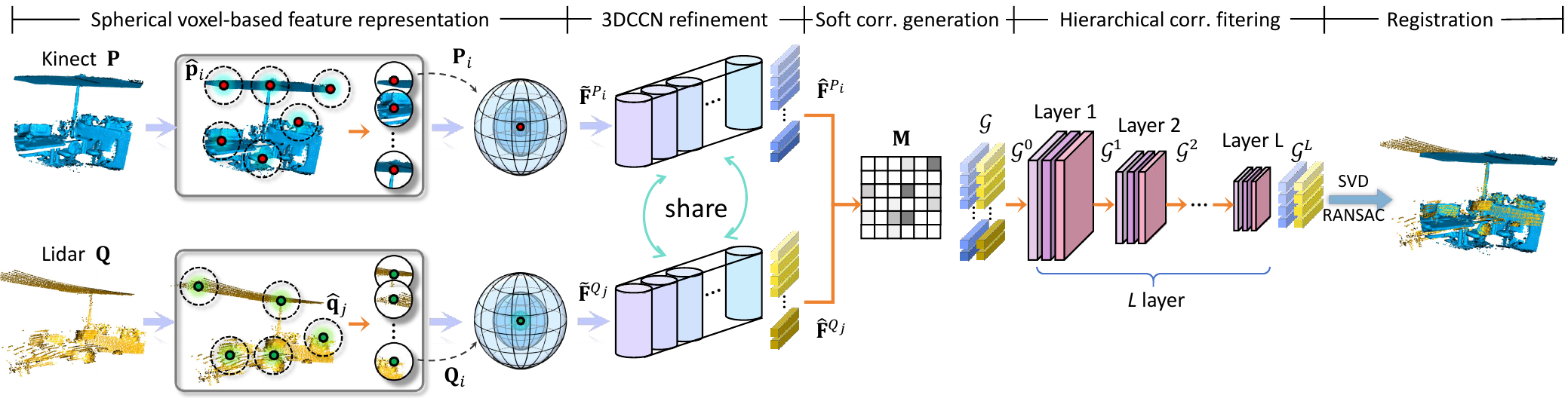}}  
  \caption{
    Given cross-source point clouds $\mathbf{P}$ and $\mathbf{Q}$, 
    we first obtain local patches $\mathbf{P}_i$ and $\mathbf{Q}_j$ by 
    FPS~\cite{qi2017pointnet} and KNN search, 
    and then complete the representation of initial 
    features $\mathbf{\widetilde{F} }^{\mathcal{P}_i}$ and 
    $\mathbf{\widetilde{F} }^{\mathcal{Q}_j}$ by spherical voxelization and multi-scale sphere normalization. 
    Then, the features are refined by the 3DCCN backbone to obtain the features 
    $  \mathbf{\widehat{F} }^{\mathcal{P}_i}$ and $  \mathbf{\widehat{F} }^{\mathcal{Q}_j}$. 
    Soft correspondence generation is applied to loosely generate the initial correspondences $\mathcal{G}$, which are further filtered 
    through hierarchical correspondence filtering to produce the final correspondence $\mathcal{G}^L$. 
    The transformation $\mathbf{T}$ is ultimately determined using SVD or RANSAC~\cite{fischler1981random}.
      }
  \label{fig1}

\end{figure*}

\section{Related Work}

\subsection{Cross-source point cloud registration}
Cross-source point cloud registration~\cite{huang2021comprehensive, huang2023cross} 
has progressed at a slower pace compared to homologous point 
cloud registration due to the inherent challenges it presents. 
The difficulty in extracting similar features for corresponding 
points in cross-source registration has limited the development 
of feature-based methods~\cite{huang2017coarse}. Instead, previous mainstream 
approaches have primarily focused on optimization-based 
methods~\cite{yang2013go, tazir2018cicp, zhao2023accurate}  
and model-based methods~\cite{huang2017coarse, huang2019fast, ling2022graph}. 
Notably, GCC~\cite{zhao2023accurate} has demonstrated superior performance 
in indoor cross-source point cloud registration, employing fuzzy clustering 
and ICP optimization~\cite{ICP}. The scarcity of cross-source data for 
training sets has hindered the development of learning-based registration methods. 
Addressing this limitation, SPEAL~\cite{xiong2023speal} has taken steps to construct a 
KITTI cross-source dataset for outdoor scenes through 
binocular estimation, enabling large-scale training.


\subsection{Correspondence fitering}
Due to challenges in cross-source data, 
feature matching is difficult, leading to the generation of n
umerous false correspondences. Therefore, correspondence filtering
plays a pivotal role in cross-source registration. 
In recent years, correspondence filtering 
methods have 
experienced rapid development, categorized into 
learning-based methods~\cite{lee2021deep, choy2020deep, PointDSC, 3DRegNet} 
and traditional geometric-based methods~\cite{leordeanu2005spectral,chen2022sc2,zhang20233d}. 
Learning-based methods~\cite{PointDSC, 3DRegNet} 
often treat correspondence filtering 
as a classification problem, eliminating correspondences 
based on class assignment. For instance, 3DRegNet~\cite{3DRegNet} incorporates 
CN-Net~\cite{CNNet} into point cloud matching, employing a confidence-based 
classification module for correspondence filtering. PointDSC~\cite{PointDSC} 
enhances the effectiveness by incorporating geometric consistency 
into the network. Non-learning methods primarily rely on determining 
geometric consistency. SC2-PCR~\cite{chen2022sc2} introduces second-order spatial 
consistency for robust correspondence filtering, demonstrating 
its efficacy. MAC~\cite{zhang20233d}  achieves optimal results by introducing the 
maximal clique concept from graph theory.

\section{Method}
\label{sec:Method}

\subsection{Spherical voxel-based feature representation}

Due to the disparate origins of point cloud data acquired from various sensors, 
there exists substantial variation in the distribution and density of cross-source 
point cloud data. This discrepancy significantly impacts feature metric learning, 
impeding the accurate determination of correspondences and resulting in registration 
failures. To address this issue, a straightforward approach involves unifying 
the densities through extensive downsampling. However, this method has the 
drawback of compromising nearest neighbor information, ultimately leading 
to feature loss. In contrast to downsampling, our paper introduces a feature 
representation based on spherical voxels to tackle the density difference 
problem. The utilization of spherical voxelization and normalization 
achieves density-invariant feature representation, followed by feature 
refinement through the 3DCCN backbone~\cite{ao2021spinnet} to 
extract patch-based features.

\ptitle{Spherical voxelization.}
We employ spherical voxelization to mitigate the density difference between the 
cross-source point clouds $\mathbf{P}$ and $\mathbf{Q}$. 
Inspired by the geographical coordinate system of the Earth (GCSE), 
we deviate from traditional cube voxelization and octree methods,
and divide the patch-based sphere along three dimensions: longitude $\theta $, latitude $\varphi$, and radius $r$, 
resulting in a voxel sphere and completing the initial feature representation. 
Notably, point cloud nearest neighbor searches involve k-nearest 
neighbor search and radius nearest neighbor search, 
with their search space approximating a sphere. 
Hence, our Spherical voxelization is better suited for 
representing nearest neighbor information 
than traditional cube voxels. Additionally, as we solely use spherical 
voxelization for feature representation, and the corresponding density-invariant 
features are normalized by the spheroid voxel without extensive downsampling, 
we incur less loss in neighbor information with the excessive downsampling.

In the original point cloud $\mathbf{P}$, we acquire $k$ keypoints $\mathbf{\widehat{P}}$
using the farthest point sampling (FPS)~\cite{qi2017pointnet}. Subsequently, 
for each $\mathbf{\widehat{p}}_i \in \mathbf{\widehat{P}}$, a radius nearest 
neighbor search is conducted within the original point cloud $\mathbf{P}$. 
The nearest neighbor points are then aggregated to form the patch $\mathbf{P}_i$
\begin{equation}
  \mathbf{P}_i =\left\{\mathbf{p}_j| \mathbf{p}_j \in \operatorname{RNN}(\mathbf{\widehat{P}}; \mathbf{\widehat{p}}_i, r)  \right\}
\end{equation}
where operator $\operatorname{RNN}$ represents a radius nearest neighbor search 
at $\mathbf{\widehat{P}}$ with $\mathbf{\widehat{p}}_i$ as the center and $r$ as the radius.
Next, we perform spherical voxelization on each patch $\mathbf{P}_i$. As 
illustrated in Figure~\ref{fig1}, we establish our patch coordinate system following GCSE. 
The spherical patch is partitioned into $N$, $M$, and $K$ regions in $\theta $, $\varphi$ , and $r$, respectively. 
Consequently, $N \times M \times K$ spherical voxels $\mathbf{V}^i_{nmk}$ with $\mathbf{\widehat{p}}_i$ as the center are generated. 
where $n$, $m$, and $k$ represent the indices of voxels 
in the three dimensions. We characterize the initial features $\mathbf{F}^{\mathcal{P}_i} \in \mathbb{R}^{N \times M \times K}$
by tallying the number of points $\mathbf{p}_i \in \mathbf{V}^i_{nmk}$ distributed 
in each small voxel $\mathbf{V}^i_{nmk}$.
In the same way, we also extract patch-based feature $\mathbf{F}^{\mathcal{Q}_i}$ 
for point cloud $\mathbf{Q}$.

\begin{figure}[htbp]
  \centering{\includegraphics[scale=0.4]{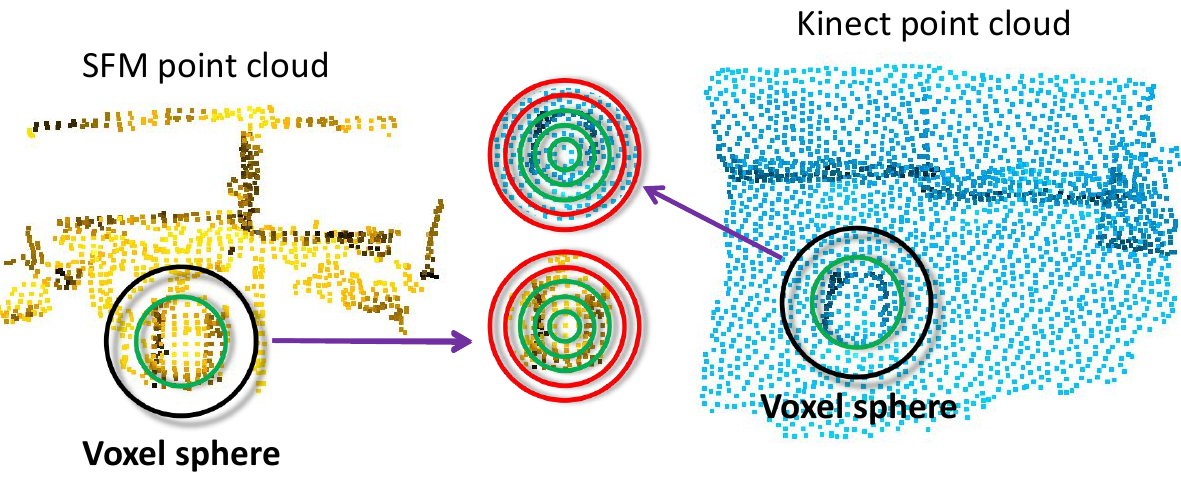}}  
  \vspace{-5pt}
  \caption{
    The superiority of our normalization method
    where the {\color{green}{green circle}} denotes the area where the 
    point distribution of the two patches is the same, and the {\color{red}{red}} indicates a difference. (For brevity, the split in the radius direction is plotted.)
    }
  \label{sphere}

\end{figure}

\ptitle{Multi-scale sphere normalization.}
We then achieve density invariance of the features by quantitatively normalizing 
the voxel spheres. The simplest normalization involves directly dividing the number of 
points in each voxel $\mathbf{V}^i_{nmk}$ of the sphere by the total number of points in the patch:
\begin{equation}
  \mathbf{\widetilde{F} }^{\mathcal{P}_i}_{nmk} = \frac{\mathbf{{F} }^{\mathcal{P}_i}_{nmk}}{\sum_{i = 1}^{N}\sum_{j = 1}^{M}\sum_{l = 1}^{K}\mathbf{{F} }^{\mathcal{P}_i}_{ijl}  }
\end{equation}
However, this normalization method has limited effectiveness for cross-source data with significant differences 
in point cloud distribution. 
Figure~\ref{sphere} demonstrates that there are differences in 
the point distribution of the outer layer of the voxel sphere.
Normalizing each layer's sphere 
separately can effectively mitigate the influence of distribution differences. 
Additionally, considering that 3DCCN~\cite{ao2021spinnet} is convolved on the sphere of each layer, 
to better suit 3DCCN, we propose multi-scale sphere normalization. Through multi-scale sphere normalization, 
we obtain the features of the $k$th layer after sphere normalization:
\begin{equation}
  \mathbf{\widetilde{F} }^{\mathcal{P}_i}_{nm\color{red}{k}} = \frac{\mathbf{{F} }^{\mathcal{P}_i}_{nm{\color{red}{k}}}}{\sum_{i = 1}^{N}\sum_{j = 1}^{M}\sum_{l = 1}^{\color{red}{k}}\mathbf{{F} }^{\mathcal{P}_i}_{ijl}  }.
\label{eq1}
\end{equation}
Using Eq.~\ref{eq1}, we apply multi-scale normalization in the radial direction 
from the outer to the inner layers. 
In comparison to normalizing the entire sphere, 
our method provides an advantage: when the point cloud distribution differs 
in the outer layer of the multi-scale sphere (Figure~\ref{sphere}), our normalization method does not affect 
the features of the inner layer. 
This enhances the robustness of our voxel representation.

\ptitle{Feature refinement.}
Although we construct the initial patch features through 
spherical voxelization and normalization, the inherent 
drawbacks~\cite{ao2021spinnet} of handcrafted descriptors, 
make these initial features susceptible to noise. This sensitivity to noise 
becomes even more pronounced in cross-source point cloud registration, 
necessitating further refinement of features to enhance robustness. Drawing 
inspiration from SpinNet~\cite{ao2021spinnet}, we refine the patch features. 
In contrast to~\cite{ao2021spinnet}, we discard point-based layers~\cite{qi2017pointnet++, ao2021spinnet} and 
leverage spherical voxel occupancy to represent the initial features. 
We then refine the feature map using 3DCCN and obtain our final 
features $\mathbf{\widehat{F} }^{\mathcal{P}_i} \in \mathbb{R}^{|\mathbf{\widehat{P}}| \times d}$
through max-pooling:
\begin{equation}
  \mathbf{\widehat{F} }^{\mathcal{P}_i} = \mathcal{M}(\operatorname{3DCCN}(\mathbf{\widetilde{F} }^{\mathcal{P}_i})), \mathbf{\widetilde{F} }^{\mathcal{P}_i}_{nmk} \in \mathbf{\widetilde{F} }^{\mathcal{P}_i}
\end{equation}
where $\mathcal{M}(\cdot)$ is max-pooling operator and $\operatorname{3DCCN}(\cdot)$ denotes performing the 3DCCN operation.

\subsection{Hierarchical correspondence filtering}
In the task of cross-source point cloud registration, the inherent distribution differences 
between point clouds pose a challenge~\cite{huang2023cross} for matching similar features at corresponding points. 
Consequently, the feature matching results exhibit a high prevalence of false correspondences.
To address this issue, filtering the 
correspondences becomes crucial. Although a straightforward solution to reduce the number of 
mismatches is to employ mutual nearest neighbor (MNN) matching~\cite{sun2021loftr}, this method is not 
suitable for the cross-source point cloud registration task. The stringent filtering conditions 
of MNN matching~\cite{sun2021loftr} may erroneously eliminate the correct 
correspondences, ultimately leading to registration failure. To overcome this challenge, 
we propose a more robust and efficient correspondence filtering method.

\ptitle{Soft correspondence generation.}
In order to capture all the correct correspondences as far as possible, 
we first conduct soft feature matching. This involves calculating the distances 
between keypoint $\mathbf{\widehat{P}}$ and $\mathbf{\widehat{Q}}$ in the feature space, yielding the matching distance 
matrix $\mathbf{S} \in \mathbb{R}^{|\mathbf{\widehat{P}}| \times |\mathbf{\widehat{Q}}|}$
\begin{equation}
\mathbf{S}_{i,j}=\frac{\mathbf{\widehat{F} }^{\mathcal{P}_i}}{(\sum_{k = 1}^{d} (\mathbf{\widehat{F} }^{\mathcal{P}_i}_k)^2 )^{1/2}}
 \cdot 
 \frac{\mathbf{\widehat{F} }^{\mathcal{Q}_j}}{(\sum_{k = 1}^{d} (\mathbf{\widehat{F} }^{\mathcal{Q}_j}_k)^2 )^{1/2}}
\end{equation}
where $d$ is the dimension of the feature and $\mathbf{\widehat{F} }^{\mathcal{Q}_j}_k$ 
denotes the $k$th value of feature $\mathbf{\widehat{F} }^{\mathcal{Q}_j}$.
The matrix $\mathbf{S}$ undergoes softmax operations 
applied to both row and column~\cite{lindenberger2023lightglue}, resulting in a soft assignment matrix $\mathbf{M}$

\begin{equation}
  \mathbf{M}_{i,j}= \underset{1 \leqslant n \leqslant |\mathbf{\widehat{P}}|}{\operatorname{Softmax}}\left(\mathbf{S}_{n j}\right)_i \underset{1 \leqslant m \leqslant |\mathbf{\widehat{Q}}|}{\operatorname{Softmax}}\left(\mathbf{S}_{i m}\right)_j
  \end{equation}
where $\operatorname{Softmax}$ denotes the softmax operation.
Utilizing the score matrix $\mathbf{M}$, 
we apply the Top-$k$ method to perform the element-wise selection, 
yielding the 
initial correspondence $\mathcal{G}$
\begin{equation}
  {\mathcal{G}}=\left\{\left(\widehat{\mathbf{p}}_{n}, \widehat{\mathbf{q}}_{m}\right) \mid\left(n, m\right) \in \operatorname{Topk}_{i,j}\left(\mathbf{M}_{i, j}\right)\right\}
\end{equation}

\ptitle{Hierarchical correspondence filtering.}
In cross-source datasets, diverse noise arises from distinct sensors~\cite{huang2023cross}. 
Simultaneously, correspondences are generated through soft matching, 
leading to a substantial proportion of outliers.  This abundance 
of outliers poses significant challenges in transformation estimation, 
rendering the robust estimator RANSAC~\cite{fischler1981random} ineffective.  
Consequently, additional filtration of false correspondences is 
imperative for achieving robust registration.  Drawing inspiration 
from SC$^2$-PCR~\cite{chen2022sc2}, we leverage the consistency of correct correspondences 
to filter the correspondences.  Aiming at the problem that the single-stage method based 
on second-order consistency~\cite{chen2022sc2} fails under a high proportion of outliers,
we introduce a progressive, hierarchical correspondence filtering method.

Taking layer $l$ as an example, the input correspondences of this filtering layer is the output of the previous layer $\mathcal{G}^{l-1}$, 
where $l \in {1, 2, \ldots, L}$ and  $\mathcal{G}^{0}=\mathcal{G}$. In Euclidean space, 
the consistency distance $d^l_{i,j}$ between correspondences $\mathcal{G}^{l-1}_i$ and $\mathcal{G}^{l-1}_j$ 
at layer $l$ is calculated:
\begin{equation}
  d^l_{i,j}= \Big| ||\mathbf{\widehat{p}}^{l-1}_i-\mathbf{\widehat{p}}^{l-1}_j||_2-||\mathbf{\widehat{q}}^{l-1}_i-\mathbf{\widehat{q}}^{l-1}_j||_2 \Big| 
\end{equation}
where $\mathcal{G}^{l-1}_i = (\mathbf{\widehat{p}}^{l-1}_i, \mathbf{\widehat{q}}^{l-1}_i) \in \mathcal{G}^{l-1}$. And the 
consistency score is defined as $s^l_{i,j}=\mathds {1} \left(d^l_{i,j}\leqslant {\sigma_d}\right)$ where $\sigma_d$ is a distance threshold.
Based on this consistency score, the second-order consistency score $ss^l_{i,j}$ between 
every two pairs of correspondences is determined:
\begin{equation}
  ss^l_{i,j}=s^l_{i,j}  \sum_{k=1}^{N^l} s^l_{i,k} \cdot s^l_{k,j}
\end{equation}
where $N^{l-1}$ is the number of correspondences $\mathcal{G}^{l-1}_i$.
Employing the second-order consistency score, 80\% of the correspondences are 
selected using the Top-$k$ (percentage) method 
as the output correspondences of $l$th layer
\begin{equation}
  {\mathcal{G}^l}=\{\mathcal{G}^{l-1}_i \mid i \in \operatorname{Topk-p}_{x}(\sum_{y = 1}^{N^{l-1}} \mathbf{ss}^l_{x, y} )\}
\end{equation}
where $\operatorname{Topk-p}_{x}(\cdot)$ represents performing Top-$k$ (percentage) operation along the $x$ axis.
As illustrated in Figure~\ref{fig1}, our final correspondences $\mathcal{G}^L$
are obtained through a hierarchical 
correspondence filtering process. This progressive
multi-step outlier removal method can robustly retain 
more correct correspondences while removing outliers, 
compared to the one-stage outlier removal methods. Finally,  
utilizing the obtained correspondences $\mathcal{G}^L$, we estimate transformation $\mathbf{T}\{\mathbf{R},\mathbf{t}\}$
through weighted SVD or RANSAC~\cite{fischler1981random}
, ensuring a robust and 
accurate cross-source registration.


\section{Experiments}
We evaluate the performance of our method on both the cross-source dataset 
3DCSR (Sec.~\ref{Cross-Source}) and the homologous dataset 3DMatch (Sec.~\ref{3DMatch}). 
In addition, we also set up ablation experiments to verify the effectiveness of 
each of our modules (Sec.~\ref{Ablation}).
The training for 3DMatch followed the experimental setup of SpinNet~\cite{ao2021spinnet}. 
However, for 3DCSR, where there is no extensive training set available, 
we address the density variations between cross-source datasets by downsampling the source 
point clouds from 3DMatch. Implementation 
details and runtime analysis are provided in Appx.~\ref{Implementation details} and \ref{results}.

\subsection{Evaluation on cross-source dataset}\label{Cross-Source}
\ptitle{Dataset.}
The 3DCSR dataset comprises 202 indoor scenes encompassing 
two cross-source types. These include point clouds transitioning 
from those collected by Kinect to those generated by Structure 
from Motion (SFM), and from Kinect to LiDAR-collected point clouds. 
The dataset consists of 37 scenes for \emph{Kinect-sfm} and 165 scenes 
for \emph{Kinect-Lidar}, both of which are classified as easy and challenging.

\ptitle{Metrics.}
Following~\cite{zhao2023accurate}, we utilize translation error (TE), 
rotation error (RE), and registration recall (RR) to evaluate our method. 
And following~\cite{qin2022geometric}, we assess correspondence results 
using inlier ratio (IR) and feature matching recall (FMR). 
Detailed metrics are explained in Appx.~\ref{Metrics}.

\begin{table}[htbp]
	\centering
    \scriptsize
	\resizebox{1.0\linewidth}{!}{
		\begin{tabular}{l|ccccc}
			\toprule
			Method & RR($\uparrow$) & IR($\uparrow$) & FMR($\uparrow$) & RRE($\downarrow$) & RTE($\downarrow$)          \\ 
            \midrule 
            \multicolumn{6}{c}{\emph{Kinect-sfm}} \\
            \midrule
            ICP~\cite{ICP} &18.8 &- &- &2.79 &0.08 	\\
            GICP~\cite{segal2009generalized} &12.5 &- &- &\textbf{1.90} &\textbf{0.03} 	\\
            JRMPC~\cite{huang2017coarse} &0 &3.3 &14.8 &- &- 	\\
            GCTR \cite{huang2019fast} &15.2 &10.9 &46.8 &2.99 &0.10 	\\
            GCC~\cite{zhao2023accurate} &81.3 &- &- &2.09 &\underline{0.06} 	\\
            SpinNet~\cite{ao2021spinnet} &\underline{84.5} &\underline{54.8} &\underline{93.8} &2.64 &0.08 	\\
            Predator~\cite{huang2021predator} &71.8 &31.8 &81.3 &3.93 &0.11 	\\
            CoFiNet \cite{yu2021cofinet} &37.5 &25.3 &90.6&4.19 &0.10 	\\
            GeoTransformer \cite{qin2022geometric} &84.4 &40.2 &90.6 &\underline{1.95} &\underline{0.06} 	\\
            VRHCF (\emph{ours}) &\textbf{93.8} &\textbf{90.3} &\textbf{96.8}  &2.06 &\underline{0.06} 	\\
            \midrule 
            \multicolumn{6}{c}{\emph{Kinect-lidar}} \\
            \midrule
 
            ICP~\cite{ICP} &1.3 &- &- &7.86 & 0.24 	\\
            GICP~\cite{segal2009generalized} &0.6 &- &- &12.8 &0.24 	\\
            JRMPC~\cite{huang2017coarse} &0 &0.5 &3.1 &- &- 	\\
            GCTR~\cite{huang2019fast} &0 &0.1 &2.1 &- &- 	\\
            GCC~\cite{zhao2023accurate} &3.1 &- &- &\underline{3.64} &\underline{0.15} 	\\
            SpinNet~\cite{ao2021spinnet} &5.2 &1.2 &7.1 &6.93 &\underline{0.15} 	\\
            Predator~\cite{huang2021predator} &5.2 &0.5 &1.3 & 4.63 &0.21 	\\
            CoFiNet \cite{yu2021cofinet} &5.2 &1.2 &7.1 &4.65 &\underline{0.15} 	\\
            GeoTransformer \cite{qin2022geometric} &\underline{9.1} &\underline{3.4} &\underline{9.7} &4.60 &0.17 	\\
            VRHCF (\emph{ours}) &\textbf{10.3} &\textbf{6.4} &\textbf{13.6}  &\textbf{3.41} &\textbf{0.13} 	\\

         \bottomrule
	\end{tabular}
    }
    \vspace{-5pt}
    \caption{
        Quantitative results on 3DCSR. 
    }
	\label{3dcsr}
\end{table}

\begin{figure}[t]
  \centering{\includegraphics[scale=0.35]{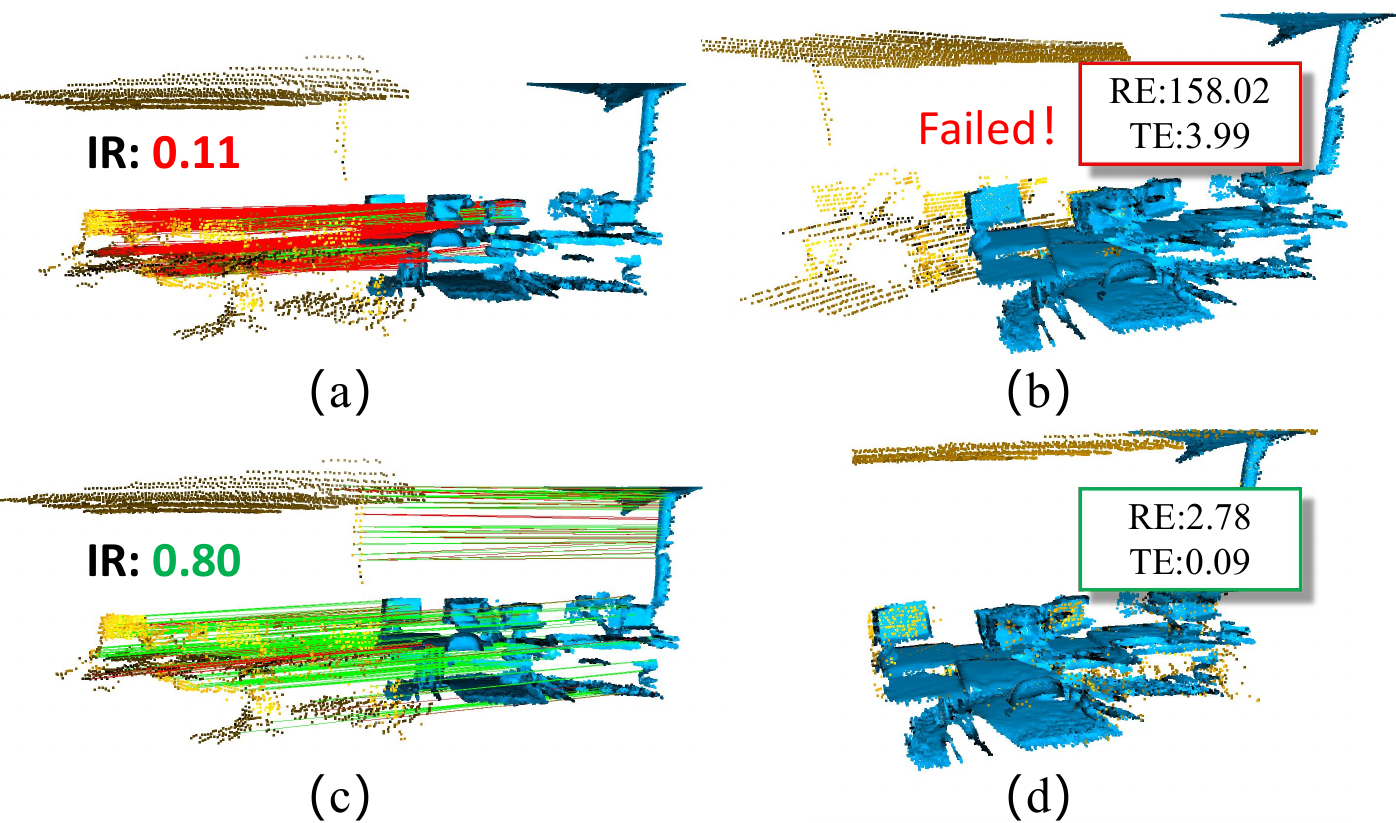}}  
  \vspace{-5pt}
  \caption{
    Qualitative results. The correct correspondences 
    are denoted by the {\color{green}{green lines}}, 
    while the {\color{red}{red lines}} indicate errors.
      }
  \label{vis}

\end{figure}

\ptitle{Results.}
We divide the experiments into two parts based on the source categories: 
\emph{Kinect-sfm} and \emph{Kinect-lidar}.  Our method is compared with 
cross-source registration methods ICP~\cite{ICP}, 
GICP~\cite{segal2009generalized}, JRMPC~\cite{huang2017coarse},
GCTR~\cite{huang2019fast}, GCC~\cite{zhao2023accurate} and 
homologous registration methods 
SpinNet~\cite{ao2021spinnet}, Predator~\cite{huang2021predator},
CoFiNet~\cite{yu2021cofinet}, GeoTrans~\cite{qin2022geometric}.
The results are presented in Table~\ref{3dcsr}.  Our method outperforms 
in 3DCSR \emph{Kinect-sfm}, with a 9.3 percentage point (pp) and 35.7 pp increase in RR and IR 
compared to the SOTA~\cite{ao2021spinnet}.  
In extremely challenging \emph{Kinect-lidar} scenarios, our method, like 
others~\cite{zhao2023accurate, qin2022geometric}, exhibits a low level in RR. 
Nevertheless, in comparison with previous methods~\cite{zhao2023accurate, qin2022geometric}, 
ours shows great improvements in all metrics, particularly with a 3.9 pp increase in FMR.
The visualization of qualitative results on 3DCSR is shown in Figure~\ref{vis}.

\subsection{Evaluation on homologous dataset}\label{3DMatch}
\ptitle{Dataset.}
The 3DMatch dataset is an indoor dataset reconstructed from RGBD images, 
comprising a total of 62 scenes. Among these, 58 scenes are allocated to the training set, 
8 scenes for the validation set, and 8 scenes for the test set. 
According to~\cite{huang2021predator}, the dataset has two benchmarks, 
distinguished by the overlap ($>$30\% or not): 3DMatch and 3DLoMatch.

\ptitle{Metrics.}
Similar to the 3DCSR experiment, we use IR, FMR, and RR to evaluate the performance. 
However, when calculating RR for homologous registration, we consider 
the registration with RMSE$<$0.2m as a successful registration.

\ptitle{Results.}
Our method is compared with 
Predator~\cite{huang2021predator},
SphereNet~\cite{SphereNet}, CoFiNet~\cite{yu2021cofinet}, 
GeoTrans~\cite{qin2022geometric} and RoITr~\cite{yu2023rotation}.
The experimental results are presented in Table~\ref{3dmatch2}. 
Our method and GeoTrans employ SVD for transformation estimation, 
while other methods utilize RANSAC~\cite{fischler1981random}. Table~\ref{3dmatch2} 
illustrates that our method attains superior results on 3DMatch, 
boasting an RR as high as 95.1\%, with a 3.2 pp increase compared with~\cite{qin2022geometric}. 
In addition, our method significantly outperforms 
in correspondence results, achieving an IR of 87.9\%, a 5.2 pp improvement over~\cite{yu2023rotation}. 
Furthermore, 
our method also excels in both RR and IR on 3DLoMatch, 
showcasing it is robust to low overlap. 
In addition, our method is also robust to the number of sampling points and achieves the best results for different numbers of sampling points.
However, our method in FMR is inferior to RoITr and GeoTrans. Because, 
in some scenarios where the original IR is very low (but IR$\geqslant $5\%), 
due to the influence of a large number of anomalies, our HCF module incorrectly 
filters out the correct correspondence, ultimately resulting in IR$<$5\%.

In summary, our method demonstrates effectiveness for both cross-source and same-source scenarios.

 

\begin{table}[htbp]
    \setlength{\tabcolsep}{2pt}
    \resizebox{1.0\linewidth}{!}{
    \centering
    \begin{tabular}{l|ccccc|ccccc}
    \toprule
     & \multicolumn{5}{c|}{3DMatch} & \multicolumn{5}{c}{3DLoMatch} \\
    \# Samples & 5000 & 2500 & 1000 & 500 & 250 & 5000 & 2500 & 1000 & 500 & 250 \\
    \midrule
    \multicolumn{11}{c}{\emph{Feature Matching Recall} (\%) $\uparrow$} \\
    \midrule
    Predator~\cite{huang2021predator} & 96.6 & 96.6 & 96.5 & 96.3 & 96.5 & 78.6 & 77.4 & 76.3 & {75.7} & 75.3 \\
    CoFiNet \cite{yu2021cofinet} & \underline{98.1} & \textbf{98.3} & \underline{98.1} & \textbf{98.2} & \textbf{98.3} & {83.1} & {83.5} & {83.3} & {83.1} & {82.6} \\ 
    GeoTransformer \cite{qin2022geometric} & 97.9 & {97.9} & {97.9} & {97.9} & {97.6} & \underline{88.3} & \underline{88.6} & \underline{88.8} & \underline{88.6} & \underline{88.3} \\ 
    SphereNet~\cite{SphereNet}  & \textbf{98.2} & \textbf{98.3} & \textbf{98.2} & 97.3 & 95.8 &79.1 &78.5&77.3&74.1&68.4 \\ 
    RoITr~\cite{yu2023rotation} &98.0  &\underline{98.0} &{97.9}  &\underline{98.0} &\underline{97.9} &\textbf{89.6}  &\textbf{89.6} &\textbf{89.5}  &\textbf{89.4} &\textbf{89.3} \\
    VRHCF (\emph{ours})  &96.7  &{96.9} &96.9  &96.9 &96.0 &82.1  &82.1 &82.0  &82.0 &81.6 \\

    \midrule
    \multicolumn{11}{c}{\emph{Inlier Ratio} (\%) $\uparrow$} \\
    \midrule
    Predator~\cite{huang2021predator} & 58.0 & 58.4 & {57.1} & {54.1} & 49.3 & {26.7} & {28.1} & {28.3} & {27.5} & 25.8 \\
    CoFiNet \cite{yu2021cofinet} & 49.8 &  51.2 & {51.9} &  52.2 &  52.2 & {24.4} & {25.9} & {26.7} & {26.8} & {26.9} \\ 
    GeoTransformer \cite{qin2022geometric} & {71.9} & {75.2} & {76.0} & {82.2} & \underline{85.1} & {43.5} & {45.3} & {46.2} & {52.9} & {57.7} \\
    SphereNet~\cite{SphereNet} & {54.2} & {54.6} & {54.3} & {48.9} & {44.2} & {24.2} & {24.8} & {25.0} & {22.7} & {19.2} \\
    RoITr~\cite{yu2023rotation} &\underline{82.6}  &\underline{82.8} &\underline{83.0}  &\underline{83.0} &{83.0} &\underline{54.3}  &\underline{54.6} &\underline{55.1}  &\underline{55.2} &\underline{55.3} \\
    VRHCF(\emph{ours})   &\textbf{87.8}  &\textbf{87.6}  &\textbf{87.6}   &\textbf{87.6}  &\textbf{87.5}  &\textbf{62.6}  &\textbf{62.6} &\textbf{62.0}  &\textbf{61.7} &\textbf{61.9} \\
    \midrule
    \multicolumn{11}{c}{\emph{Registration Recall} (\%) $\uparrow$} \\
    \midrule
    Predator~\cite{huang2021predator} & 89.0 & 89.9 & {90.6} & 88.5 & 86.6 & 59.8 & 61.2 & 62.4 & 60.8 & 58.1 \\
    CoFiNet \cite{yu2021cofinet} & 89.1 & 88.9 & {88.4} & {87.4} & {87.0} & {67.5} & {66.2} & {64.2} & {63.1} & {61.0} \\ 
    GeoTransformer \cite{qin2022geometric} & \underline{92.0} & \underline{91.8} & \underline{91.8} & \underline{91.4} & \underline{91.2} & \underline{75.0} & \underline{74.8} & {74.2} & {74.1} & {73.5} \\
    SphereNet~\cite{SphereNet}  & 91.2 & 90.3 & 88.8 & 86.2 & 77.7  &60.0 & 59.6&53.9&45.5&32.4 \\
    RoITr~\cite{yu2023rotation}  &91.9  &91.7 &\underline{91.8}  &\underline{91.4} &91.0  &{74.7}  &\underline{74.8} &\underline{74.8}  &\underline{74.2} &\underline{73.6} \\
    VRHCF (\emph{ours}) &\textbf{95.1}  &\textbf{94.6} &\textbf{94.9}  &\textbf{94.9} &\textbf{94.3}  &\textbf{76.8}  &\textbf{76.8} &\textbf{76.7}  &\textbf{76.7} &\textbf{76.4} \\
    \bottomrule
    \end{tabular}}
    \vspace{-5pt}
    \caption{
    Quantitative results on 3DMatch and 3DLoMatch.
    }
    \label{3dmatch2}
    \vspace{-10pt}
\end{table}

\begin{table}[t]
  \scriptsize
  \setlength{\tabcolsep}{1.5pt}
  \centering
  \resizebox{1.0\linewidth}{!}{
  \begin{tabular}{l|ccccc|ccc|ccc}
  \toprule
  & \multicolumn{5}{c|}{}&\multicolumn{3}{c|}{{3DCSR}} & \multicolumn{3}{c}{{3DMatch}} \\
  No. &SV &MSN&SCG &MNN &HCF           & RR($\uparrow$)  & IR($\downarrow$) & FMR($\downarrow$) & RR($\uparrow$)  & IR($\downarrow$) & FMR($\downarrow$) \\
  \midrule

  1)             &\checkmark    &\checkmark   &\checkmark  & &\checkmark  &\textbf{93.8}    &\textbf{90.3}    &\textbf{96.8}   &\textbf{95.1} &{87.8} &{96.7}  \\
  2)             &\checkmark    &   &\checkmark  & &\checkmark                      & 87.8      & 86.8      & 90.1       & {94.9}    & {87.4}    & 96.3    \\
  3)             &              &   &\checkmark  & &\checkmark                    & 84.5      & 82.4       & 93.8    &93.0     &85.2  &96.3       \\
  4)             &\checkmark    &\checkmark   & &\checkmark  &\checkmark            &84.5       &89.7       &87.8        &\textbf{95.1}     &\textbf{88.2}     &96.2   \\
  5)             &\checkmark    &\checkmark   &\checkmark  & &           &90.1   &57.4  &90.6  &91.0   &52.1  &\textbf{97.9}   \\

  \bottomrule
  \end{tabular}
  }
  \vspace{-5pt}
  \caption{
     Ablation results on 3DCSR and 3DMatch.
     }
  \label{tab:Ablation}
\end{table}

\subsection{Ablation study}\label{Ablation}
We conduct ablation experiments to assess the effectiveness of each module. 
We present a complete model and four ablation models, with the experimental results detailed in Table~\ref{tab:Ablation}.


\ptitle{Multi-scale sphere normalization (MSN).}
The model without MSN experiences a notable performance 
decline on the 3DCSR dataset, as depicted in (1) and (2) of Table~\ref{tab:Ablation}. 
This emphasizes the significance of our MSN module, which 
addresses the density inconsistency issue and is effective for 
cross-source registration.

\ptitle{Spherical voxelization (SV).}
As depicted in (1) and (3), the performance of the model without SV
further decreases  on both datasets. 
Because SV plays a pivotal role in feature representation
and contributes to achieving robust registration.

\ptitle{Soft correspondence generation (SCG) or mutual nearest neighbor (MNN).}
Shown in (1) and (4), the model with SCG outperforms the model with MNN on 3DCSR, 
while its performance degrades on 3DMatch. This discrepancy arises 
while its performance degrades on 3DMatch. This discrepancy arises 
because the SCG module is more suitable for scenarios with numerous 
false correspondences, demonstrating superior performance in cross-source registration.

\ptitle{Hierarchical correspondence filtering (HCF).}
Illustrated in (1) and (5), the incorporation of the HCF module significantly 
enhances the performance of our method compared to the case without the HCF module. 
It plays a crucial role in achieving robust and accurate registration.

\section{Conclusion}
This paper introduces a novel framework for cross-source point cloud registration, 
aiming to realize robust and accurate registration. To address challenges arising 
from inconsistent density and distribution differences 
in cross-source data, we propose a spherical voxel-based feature representation for feature extraction. 
Additionally, to tackle difficulties in matching cross-source point features 
and mitigating the impact of numerous mismatches, 
we present soft correspondence generation and hierarchical correspondence filtering. 
These methods aim to enhance the quality of correspondences, 
leading to more robust registration. 
In the future, an indoor cross-source point cloud dataset will be collected and produced for large-scale training.

\bibliographystyle{IEEEbib}
\bibliography{icme2022template}

\appendix
\clearpage

\section{Implementation details}\label{Implementation details}

Our approach is implemented using PyTorch and Open3D. We conducted two 
distinct preprocesses on the 3DMatch dataset to generate training sets 
for both the homologous registration model and the cross-source registration model. 
For homologous registration, following~\cite{huang2021predator}, we downsample 
the point cloud data using 0.02m voxels. To simulate density inconsistency in 
cross-source point cloud data for cross-source registration, 
we further processed 3DMatch. The source and target point clouds underwent 
data augmentation through random voxel downsampling, with voxel side lengths following 
a uniform distribution $\mathbf{U}(0,0.1)$ (it's worth noting that 
the voxel side lengths for the source and target point clouds are uncorrelated). 
During training, we utilize a batch size of 64, and the initial 
learning rate is set to 0.001, reducing by 50\% every 5 epochs. 
All experiments reported in this paper are conducted on a single Nvidia 
RTX 3090 with 24GB memory.

\section{Detailed Metrics}\label{Metrics}
Following~\cite{zhao2023accurate}, we utilize Translation Error (TE), 
Rotation Error (RE), and Registration Recall (RR) to evaluate our method. 
Additionally, Following~\cite{qin2022geometric}, we assess correspondence results 
using Inlier Ratio (IR) and Feature Matching Recall (FMR). 
(1) RE: the geodesic distance between the ground-truth 
and the estimated rotation matrix.
(2) TE: the Euclidean distance between the ground-truth and the estimated translation vector. 
(3) RR: the proportion of registered successfully point cloud pairs (RE$<$15$^{\circ}$ and TE$<$30cm) 
to the total point cloud pairs in the dataset. 
(4) IR: the proportion of estimated correct correspondences ($<$0.1m) to the total ground-truth correspondences.
(5) FMR: the proportion of point clouds with successful feature matching (IR$>$0.05) to the total point cloud pairs in the dataset.

\ptitle{Inlier Ratio (IR)} is the proportion of estimated correct 
correspondences ($<$0.1m) to the total ground-truth correspondences:
\begin{equation}
  \mathrm{IR}_h=\frac{1}{\left|\mathcal{G}_h\right|} \sum_{\left(\mathbf{p}_i, \mathbf{q}_j\right) \in \mathcal{G}_h} \mathds{1}\left(\left\|\mathbf{T}_h(\mathbf{p}_i)-\mathbf{q}_j\right\|_2<\tau_1\right)
\end{equation}
where $\mathcal{G}_h$ is the correspondences of the $h^{th}$ point cloud pair in the test dataset and 
$\mathds{1}$ is an indicator function. $\mathbf{T}(\mathbf{p}_i)$ is the point after 
transformation $\mathbf{T}$ from $\mathbf{p}_i$.

\ptitle{Feature Matching Recall (FMR)} represents the proportion 
of point clouds with successful feature matching ($  \mathrm{IR}_h>$0.05) 
to the total point cloud pairs in the dataset
\begin{equation}
  \mathrm{FMR}=\frac{1}{H} \sum_{h=1}^H \mathds{1}\left(\mathrm{IR}_h>\tau_2\right)
\end{equation}
where $H$ is the number of point cloud pairs in the dataset.

\ptitle{Rotation Error (RE)} represents the geodesic distance between the ground-truth rotation matrix $\mathbf{R}_h$
and the estimated rotation matrix $\hat{\mathbf{R}}_h$:
\begin{equation}
  \mathrm{RE}_h=\arccos \left(\frac{\operatorname{trace}\left(\hat{\mathbf{R}}_h^T \mathbf{R}_h\right)-1}{2}\right)
  \end{equation}

\ptitle{Translation Error (TE)} represents the Euclidean distance between the ground-truth  translation vector
$\mathbf{t}$: and the estimated translation vector $\hat{\mathbf{t}}_h$:
\begin{equation}
  \mathrm{TE}_h=\left\|\hat{\mathbf{t}}_h-\mathbf{t}_h\right\|
\end{equation}

\ptitle{Registration Recall (RR)}
Following~\cite{zhao2023accurate, huang2017coarse}, 
for cross-source registration, $\mathrm{RE}_h<15^{\circ}$ and $\mathrm{TE}_h<30$cm is 
considered as successful registration.
Therefore, RR in the cross-source point cloud registration can be calculated as:
\begin{equation}
    \mathrm{RR}=\frac{1}{H} \sum_{h=1}^H \mathds{1}\left(\mathrm{TE}_h<0.3 \mathrm{~m} \wedge  \mathrm{RE}_h<15^{\circ}\right)
\end{equation}

\section{More experimental results}\label{results}

\subsection{More qualitative results}
\begin{figure*}[t]
  \centering{\includegraphics[scale=0.8]{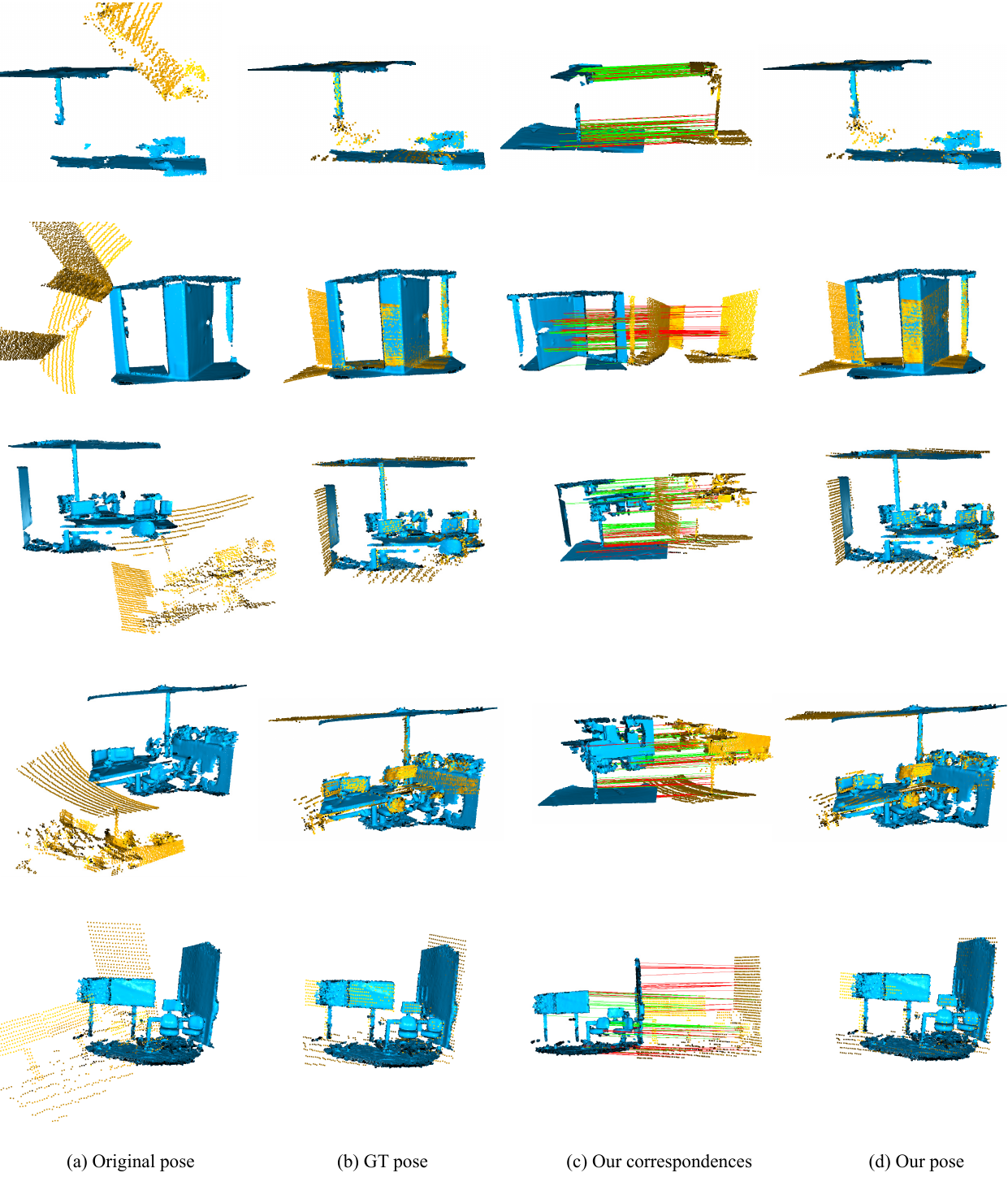}}  
  \caption{
    qualitative results on Kinect-SFM.
      }
  \label{qr1}
\end{figure*}

\begin{figure*}[t]
  \centering{\includegraphics[scale=0.8]{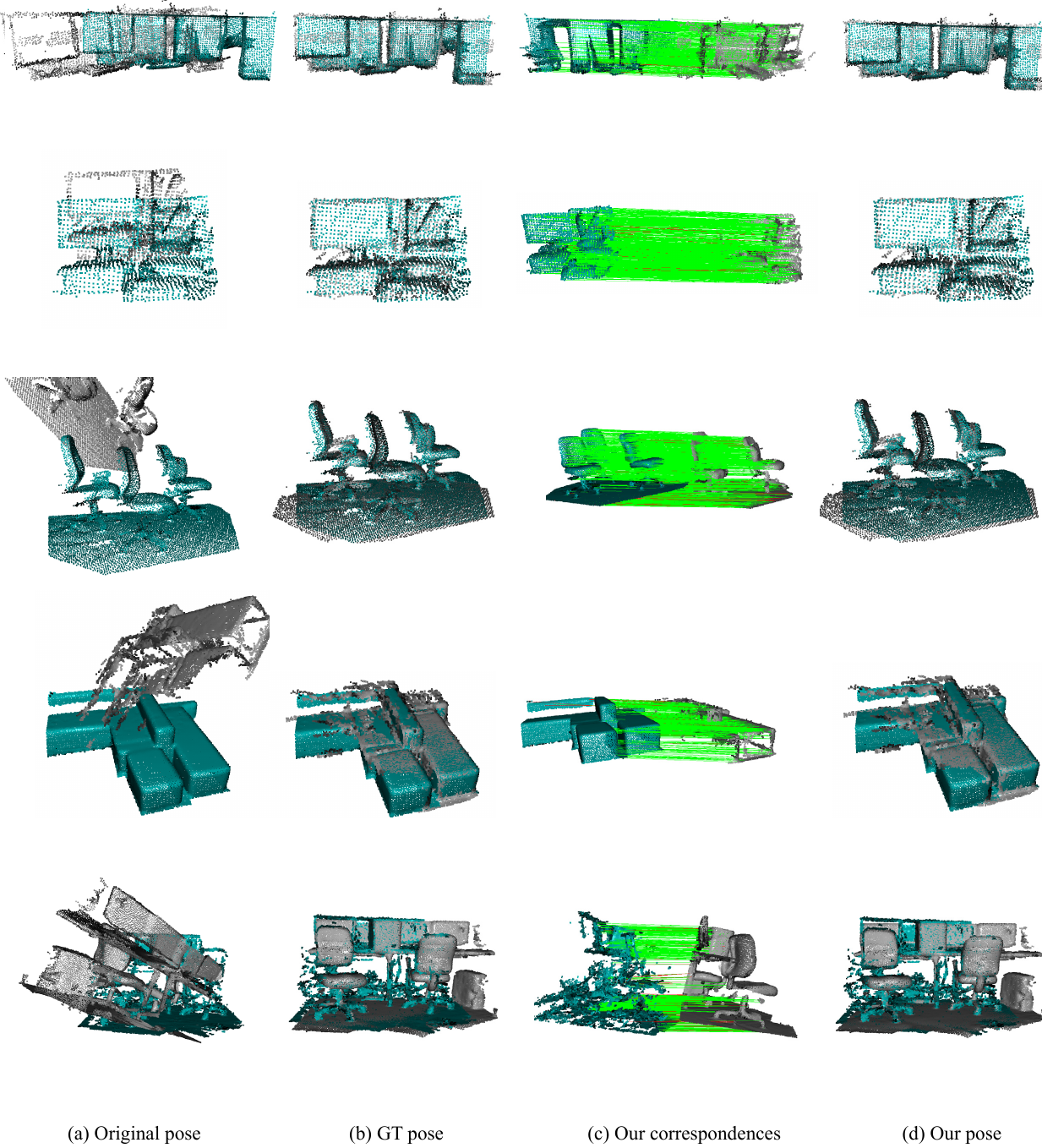}}  
  \caption{
    qualitative results on Kinect-SFM.
      }
  \label{qr2}
\end{figure*}

In this section, we present additional visualization 
experiments and qualitative results. As illustrated 
in Figure~\ref{qr1} and~\ref{qr2}, our method demonstrates robust and accurate 
registration with low RE and TE. 
Moreover, owing to our hierarchical correspondence 
filtering, our method obtains high-quality correspondences.

\subsection{Runtime analysis}

\begin{table}[htbp]
	\centering
	\resizebox{1.0\linewidth}{!}{
		\begin{tabular}{l|ccc|ccc}
			\toprule
            &\multicolumn{3}{c|}{\emph{Kinect\_sfm}} & \multicolumn{3}{c}{\emph{Kinect\_lidar}} \\
			      Method & Feature (s) $\downarrow$& Pose (s) $\downarrow$ & Total (s) $\downarrow$   & Feature (s) $\downarrow$& Pose (s) $\downarrow$ & Total (s) $\downarrow$      \\ 
      \midrule  
            ICP~\cite{ICP} &- &\underline{0.83} &\underline{0.83}	&- &4.03 &4.03\\
            RANSAC~\cite{fischler1981random} &41.9 &- &- &4.10 &0.14 	\\
            GCC~\cite{zhao2023accurate} &- &2.98 &2.98  &- &\underline{4.21} &\underline{4.21}\\
            \midrule        
            Predator~\cite{huang2021predator} &{0.10} &0.18 & 0.28	 &0.68 &0.25 &0.93\\
            CoFiNet \cite{yu2021cofinet} &0.31 &0.28 &0.59	 &0.19 &0.22 &0.41\\
            GeoTrans \cite{qin2022geometric} &\textbf{0.07} &\textbf{0.06} &\textbf{0.13}  &\textbf{0.09} &\underline{0.14} &\textbf{0.23}	\\
            \midrule  
            Perfectmatch~\cite{gojcic2019perfect} &76.31 &0.88 & 77.19	 &108.90 &0.97 &109.87\\
            SpinNet~\cite{ao2021spinnet} &106.29 &1.50 & 107.79	 &159.49 &1.16 &160.65\\
            VRHCF (\emph{ours}) &\underline{8.27} &\underline{0.10} &\underline{8.32}  &\underline{8.38} &\textbf{0.12} &\underline{8.60} \\
      \bottomrule
	\end{tabular}
    }
    \caption{
        Runtime analysis on 3DCSR.
    }
	\label{Runtime}
\end{table}

We conduct experiments to assess the running time, 
and the results are presented in Table~\ref{Runtime}. 
The first three in Table~\ref{Runtime} are traditional algorithms, 
the middle three are fragment-based methods, and the last three are patch-based methods.
Although our method belonging to patch-based methods is far inferior to fragment-based methods in speed,
our method exhibits 
the best performance in runtime efficiency compared 
with other patch-based methods~\cite{ao2021spinnet, gojcic2019perfect}. 
Regarding feature extraction, 
although we choose the time-consuming 3DCCN backbone for 
the consideration of generalization performance, we abandon 
the point layer and replace it with simple and efficient spherical voxelization, 
which greatly reduces the time consumption of our feature extraction. 
In solving the pose, although we use multi-layer correspondence filtering, 
it only performs Topk selection based on the consistency score. 
Finally, we use SVD to solve the pose, which is much faster than RANSAC.

\end{document}